# Integration of spatio-temporal contrast sensitivity with a multi-slice channelized Hotelling observer

Ali N. Avanaki[a1], Kathryn S. Espig[a], Cedric Marchessoux[b], Elizabeth A. Krupinski[c], Predrag R. Bakic[d], Tom R. L. Kimpe[b], Andrew D. A. Maidment[d]

[a] Barco Healthcare, Beaverton OR; [b] Barco Healthcare, Kortrijk, Belgium;
[c] Departments of Radiology and Psychology, University of Arizona, Tucson AZ;
[d] Department of Radiology, University of Pennsylvania, Philadelphia PA

**ABSTRACT**

Barten's model of spatio-temporal contrast sensitivity function of human visual system is embedded in a multi-slice channelized Hotelling observer. This is done by 3D filtering of the stack of images with the spatio-temporal contrast sensitivity function and feeding the result (i.e., the *perceived* image stack) to the multi-slice channelized Hotelling observer. The proposed procedure of considering spatio-temporal contrast sensitivity function is generic in the sense that it can be used with observers other than multi-slice channelized Hotelling observer. Detection performance of the new observer in digital breast tomosynthesis is measured in a variety of browsing speeds, at two spatial sampling rates, using computer simulations. Our results show a peak in detection performance in mid browsing speeds. We compare our results to those of a human observer study reported earlier (I. Diaz et al. SPIE MI 2011). The effects of display luminance, contrast and spatial sampling rate, with and without considering foveal vision, are also studied. Reported simulations are conducted with real digital breast tomosynthesis image stacks, as well as stacks from an anthropomorphic software breast phantom (P. Bakic et al. Med Phys. 2011). Lesion cases are simulated by inserting single micro-calcifications or masses. Limitations of our methods and ways to improve them are discussed.

## 1. INTRODUCTION

The human visual system (HVS) does not respond equally to excitations with different spatial or temporal frequencies. More specifically, the just-noticeable visibility threshold varies for contrast excitations with different frequencies. The inverse of contrast visibility threshold function is called contrast sensitivity function (CSF) [1, 4]. Figure 1 depicts a typical spatial-temporal CSF (stCSF) vs. temporal and spatial frequencies of the excitation.

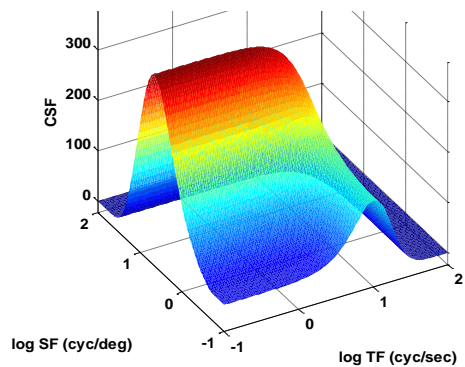

**Figure 1.** Spatio-temporal CSF [1] for 2.5° field @ 20 cd/m$^2$. Other parameters are listed in Section 2.2. SF and TF are spatial and temporal frequencies of the excitation.

---

[1] Corresponding author

In this work, we add Barten's stCSF to a model observer. Barten model is selected because it can be dynamically adjusted to the viewing conditions and is also used in developing DICOM GSDF. Our goal is to enhance the model observer by making it perform more similarly to a human observer. Doing so is especially important in observer modeling for modalities that produce a time-varying or three-dimensional output such as digital breast tomosynthesis, where the images are viewed by sequentially browsing through the stack. Note that the proposed procedure of applying stCSF is generic and may be used along with any model observers to factor in this effect.

### 1.1 Prior work

In [2], Barten's spatial CSF (ch. 3 of [1]) is integrated with a channelized model observer. Then, in application of CSF, to calculate the perceived (contrast-thresholded) images which are fed to the observer, the image is assumed to have a single spatial frequency given by Eq. (18) of [2]. A single free parameter, $C$ in Eq. (17) of [2] that controls the thresholding level, is adjusted to best match the results of a human observer study for a variety of signal strengths.

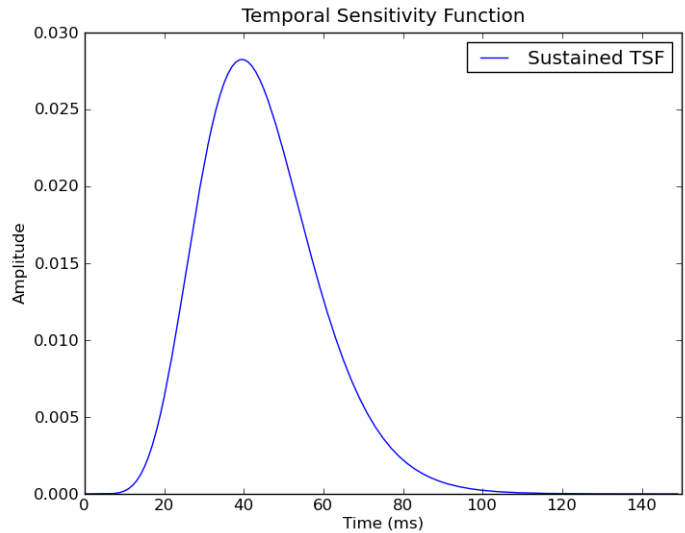

**Figure 2.** "Sustained temporal sensitivity function," from Figure 4 of [3], describes limitation of HVS response to excitations in time.

To the best of our knowledge, HVS's temporal CSF has not been used in conjunction with any linear model observer yet. In [3], only time limitation of HVS response to excitations (Figure 2) is considered. Doing so ignores the dependence of temporal CSF on the spatial frequency content of the excitation. As an example (see Figure 1), the temporal CSF is low-pass for medium and high (1 cyc/deg or higher) spatial frequencies, but it is band-pass for lower spatial frequencies (e.g. 0.1 cyc/deg).

## 2. METHODS AND MATERIALS

A block diagram of our simulations is shown in Figure 3. Simulations are performed in MEVIC (Medical Virtual Imaging Chain) [7], an extensible C++ platform developed for medical image processing and visualization. The perceived sequence of images is found by application of stCSF to the image stack at the output of display simulation. Further detail is given in Section 2.2. The splitter divides the healthy or lesion stacks to $n + 1$ non-overlapping subsets, where $n$ is the number of readers (i.e., identical msCHOs). Each reader will use one lesion subset and one healthy subset for training (i.e., calculation of weights in the thresholding equation; see [8] for details). Then all readers use the same subset of lesion stacks and the same subset of healthy stacks for testing. The scores each reader produced for each tested stack are then aggregated by one-shot Multi-Reader Multi-Case (MRMC; [9]) analysis to produce the average AUC (Area Under Curve; [19, 20, 21]) and its variance. The number of readers, $n$, may be empirically adjusted to minimize

AUC variance. We used $n = 5$ for 6000-stack real DBT set and $n = 4$ for the set with 3296 stacks from software breast phantom (see Section 2.5 for more details on the datasets).

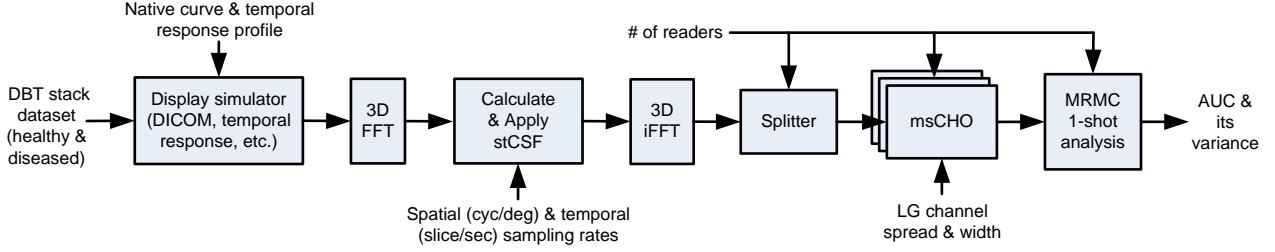

**Figure 3.** Block diagram of our simulations. 3D FFT & iFFT transform the stack to/from spatio-temporal frequency domain, where CSF is applied.

### 2.1 Display simulation

BARCO MDMG 5221 medical display ($L_{min}$ = 1.05 cd/m², $L_{max}$ = 1000 cd/m²), which is optimized and approved by FDA for reading of digital breast tomosynthesis images with RapidFrame temporal response compensation technology [13], is used for simulations.

### 2.2 stCSF model

Barten's stCSF model, Eq. (5.2) in [1], is given by

$$S(u,w) = \frac{M_{opt}(u)}{k \sqrt{\frac{2}{T}\left(\frac{1}{X_0^2} + \frac{1}{X_{max}^2} + \frac{u^2}{N_{max}^2}\right)\left(\frac{1}{\eta p E} + \frac{\Phi_0}{[H_1(w)\{1 - H_2(w)F(u)\}]^2}\right)}} \qquad (1)$$

in which $u$ and $w$ are spatial (in cyc/deg) and temporal (in cyc/sec) frequencies respectively. For completeness, all terms of (1) are given below. From Eq (3.21) in [1], we have

$$F(u) = 1 - \sqrt{1 - \exp(-(u/u_0)^2)} \qquad (2)$$

with $u_0 = 7$. For $H_i(w)$, $i = 1, 2$, we use Eq. (5.5) in [1]:

$$H_i(w) = \sqrt{\left(1 + (2\pi\tau_i w)^2\right)^{-n_i}}, \quad i = 1, 2 \qquad (3)$$

with parameters given by Eqs. (5.11) and (5.12) in [1]:

$$\tau_1 = \frac{\tau_{10}}{1 + 0.55\ln\left\{1 + (1+D)^{0.6}\frac{E}{3.5}\right\}}, \qquad (4)$$

$$\tau_2 = \frac{\tau_{20}}{1 + 0.37\ln\left\{1 + \left(1 + \frac{D}{3.2}\right)^5 \frac{E}{120}\right\}} \qquad (5)$$

where $n_1 = 7, n_2 = 4, \tau_{10} = 32e-3, \tau_{20} = 18e-3, D = 2X_0/\sqrt{\pi}$ are set per recommendations in [1]. Also,

$$M_{opt}(u) = \exp(-2(\pi\sigma u)^2) \tag{6}$$

$$\sigma = \frac{1}{60}\sqrt{\sigma_0^2 + (C_{ab}d)^2} \tag{7}$$

$$E = \frac{\pi d^2 L}{4}\left(1 - (d/9.7)^2 + (d/12.4)^4\right) \tag{8}$$

$$d = 5 - 3\tanh\{0.4\ln(LX_0^2/40^2)\} \tag{9}$$

in which $L$ is the average luminance of the object (over space & time, in cd/m$^2$), $X_0$ is the apparent size of the image (in degrees; depends on image size in pixels, pixel pitch, and the viewing distance, assuming a viewing axis orthogonal to the image plane), $d$ is pupil diameter in mm, and $E$ is proportional to retinal illuminance, in Trolands. $M_{opt}$ is the optical modulation transfer function of the eye, and $\sigma$ is the standard deviation of the line-spread function, with $C_{ab} = 0.08$ and $\sigma_0 = 0.5$. The rest of parameters is $\eta = 0.03$, $\Phi_0 = 3e-8$, $X_{max} = 12$, $N_{max} = 15$, $T = 0.1$, $p = 1.285e6$ (from Table 3.2 in [1]) and $k = 3$.

Note that instead of (3), if we substitute $H_i(w) = 1$ in (1), $S(u,w)$ becomes only a function of $u$ and is reduced to the spatial-only CSF given by Eq. (3.26) of [1] that is used in [2].

Our implementation is validated against the graphs in Figures 3.5, 5.2, and 5.3 of [1].

### 2.3 Applying stCSF

Our method for applying the stCSF to the input stacks is based on the following two propositions.

**Proposition 1.** Barten's stCSF model includes only a single spatial frequency. Our data includes 2 spatial frequencies, $u_1$ and $u_2$, which we combine to make a single spatial frequency as follows:

$$u = \sqrt{u_1^2 + u_2^2} \tag{10}$$

**Proposition 2.** For simplicity, we assume that each spatio-temporal frequency component of the input (i.e., displayed DBT stack) can be processed independently from others. This is equivalent to ignoring the contrast masking effect (see Section 5). From the definition of CSF, we have:

$$\frac{1}{S(u,w)} = \frac{a_{JN}(u,w)}{L} \rightarrow a_{JN}(u,w) = \frac{L}{S(u,w)} \tag{11}$$

where $a_{JN}(u,w)$ is the just-noticeable amplitude of the component with the spatio-temporal frequency of $(u, w)$, and $u$ is substituted from (10). $S(u, w)$ is given by (1) and $L$ is defined in Section 2.2. The amplitude of the component, $a(u, w)$, is therefore, in a first order approximation, translated to the component's *perceived* amplitude of

$$\frac{a(u,w)}{a_{JN}(u,w)} = a(u,w)\frac{S(u,w)}{L} \tag{12}$$

in JNDs (Just Noticeable Difference).

**Application.** To apply stCSF, we decompose each image stack contrast (i.e., stack luminance values – $L$) into its spatio-temporal frequency components using a 3D fast Fourier transform (FFT). The perceived amplitude of each component is then calculated using (12). Finally, an inverse 3D FFT (iFFT) of this yields the perceived stack (i.e., with stCSF accounted for) which is fed to the model observer.

Some practical considerations in application of stCSF follow.
- Double precision FFT/iFFT is used since single precision produced large round-off errors in a typical simulation.
- We use FFTW library [16] which only produces the result array. The frequency $(u_1, u_2, w)$ for each element of the array is needed to calculate $S$ in (12). We have (see [10]):

$$u_1 = \begin{cases} \dfrac{k}{N} f_s & k < \dfrac{N}{2} \\ \left(\dfrac{k}{N} - 1\right) f_s & \text{o.w.} \end{cases}, \quad k = 0, 1, ..., N-1 \tag{13}$$

  where $k$ is the element's index on the dimension corresponding to $u_1$, $N$ is the array size on that dimension, and $f_s$ is the sampling rate for the dimension. $u_2$ and $w$ are calculated similarly.
- The minimum valid value of $u$ is $u_0 = (2X_0)^{-1}$ (p. 98 of [1]). Per Barten's suggestion, we replace $S(u, w)$ with $S(u_0, w)$ for $u > u_0$.
- FFTW is not unitary. To be in JNDs, the final result array should be divided by stack size (image width × height × number of slices) before passing it to the readers.
- Barten does not mention a phase for stCSF. The procedure described above is a zero-phase filtering. Hence, application of stCSF does not change the position of the artifact that is spatio-temporally centered in the lesion stacks of the input dataset.
- To alleviate artifacts caused by Fourier analysis, margins of all images (64×64 in our datasets) in stacks were tapered to zero starting from five pixels away from the margin. Tapering in time is not required because when viewed by a human observer the stacks are commonly browsed repeatedly (i.e., may be assumed periodic in time).

### 2.4 Model observer

We use type 'b' msCHO (see the middle diagram of Figure 6 in [8]). In this model, the central slice of the stack is used for training a 2D CHO which is applied to a set of slices around (and including) the central slice in the testing phase (i.e., "reading"). We used 15 LG channels at a spread of 10. The set of central slices for testing only include those slices that are affected in lesion insertion process.

### 2.5 Datasets

The following datasets are used in the simulations.

**Dataset A**. This set consists of 6000 reconstructed DBT stacks extracted from from clinical data; the same data was used in [15]. Half of stacks are labeled as "healthy," (i.e., without lesions) and the other half, labeled "lesion," have a single micro-calcification of selected density inserted in the stack's spatio-temporal center. Each stack is comprised of 41 slices with 1mm slice separation. Each slice is a 10-bit 64×64 image. The insertion of single microcalcifications consisted of warping 2D lesions extracted from clinical data (provided by the University of Arizona), followed by an interpolation between DBT slices [15].

**Dataset B.** This set consists of 3296 stacks from an anthropomorphic software breast phantom developed at the University of Pennsylvania [18]. The phantom simulates the breast anatomy based on an analysis of histological and radiological images. The arrangement of breast tissues at the large and medium spatial scales is realistically simulated using a region growing approach [22]. First, 1648 stacks are manually selected and cropped from the phantom while trying to evenly represent all background complexity levels. Each stack is comprised of 32, 10-bit 64×64 slices, with 200 micron slice separation. Five slices of a sample stack are shown in Figure 4. Lesion stacks are generated from the healthy ones by insertion of masses or micro-calcifications in the center of the stack using the same procedure as in Dataset A. Before being fed to an observer, the stacks are ordered to ensure that no healthy stack and its corresponding lesion version fall into the same training or test sets.

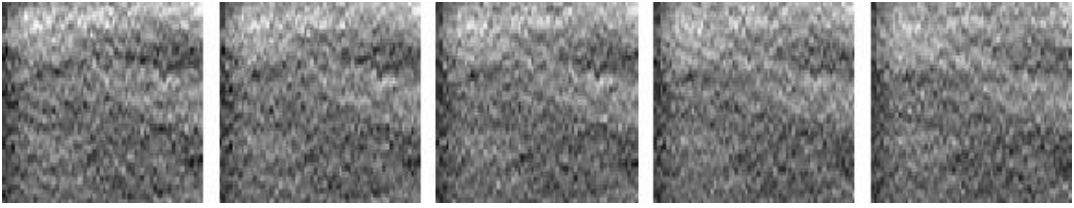

**Figure 4.** Slices 14 to 18 of a sample stack in dataset B.

### 2.6 Foveal vision

In a fixed-eye viewing scenario (e.g., when the observer knows that the lesion, if present, is at the center of the image), it is important to note that the acuity of human vision is not the same across the field of view [6, 17]. We model this effect at the end of the stCSF pipeline as follows. The variable that controls foveal vision in the simulation can take one of the following three states.

- None: means no foveal vision simulation.
- Hard thresholding: after stCSF, pixels that are away 7 degrees or more from the viewing axis are forced to zero.
- Soft thresholding: after stCSF, based on its angle from the viewing axis, α, each pixel is weighted with the coefficient $C_{fv}(\alpha)$ given by

$$C_{fv}(\alpha) = \begin{cases} 0.02 & 63.5780 < \alpha \\ -\sum_{i=0}^{6} b_i q^i & \text{o.w.} \end{cases}, \quad \begin{aligned} b_0 &= 0.04526296245190 & b_1 &= 4.48579690404659 \\ b_2 &= 21.9046292071393 & b_3 &= 55.8322547230034 \\ b_4 &= 58.6385398078192 & b_5 &= 19.7119376682204 \\ b_6 &= 1.43849397325222 & q &= -(\alpha + 0.1)^{-1} \end{aligned} \quad (14)$$

which is a tight polynomial fit to the relative acuity curve in [6].

## 3. RESULTS AND DISCUSSION

The goal of the simulations reported in the following is to find a set of viewing conditions optimum for detection of artifacts.

### 3.1 Browsing speed

For dataset A (described in Section 2.5), the detection performance for a variety of browsing speeds with spatial sampling rate (SSR) fixed at 7 or 14 pixels/deg is listed in Table 1. To compare our results with the human observer results reported in Figure 5 of [3] (two $P_{Detection}$ vs. browsing speed graphs, for micro-calcification and mass lesions respectively), we overlay their data on ours (Figure 5), after adjusting the mean and variance of their data to ours as follows: $P_{Detection}$ values which are only available at browsing speeds 5, 15, and 30 slice/sec, are changed to have the same average and standard deviation as those of our (AUC) results for the same browsing speeds; the tolerance values associated with the $P_{Detection}$ values (i.e., error bar lengths) are scaled similarly. Doing so is justified because we are not using the same dataset as the one used in [3], and AUC may be considered the average of $P_{Detection}$ over $P_{False\ alarm}$.

In [14] (see Table 1 therein), AUC was constant at 0.800±0.014 for a wide range of browsing speeds. In [15], the detection performance increases with lowering browsing speed (see Figure 6 therein: higher frame repeat means lower browsing speed). This is also the case for both CHO and PPW (partial pre-whitening) model observers used in Table 3 of [3] (micro-calcification rows). None of these results can explain the trend in human observer study (more specifically, the drop in AUC for browsing speed of 5 slice/sec). This is perhaps because (i) The model observers in [14, 15] did not consider CSF. (ii) A low-pass model for temporal contrast sensitivity, independent from spatial frequency, is used by the model observers in [3]. Barten's CSF model is, however, a spatio-temporal one and is better approximated by a band-pass behavior for a variety of frequencies. We did not find enough information in [3] to approximate the SSR used in their experiment. From Figure 5, it may be observed that our model better fits the human observer results at SSR of 7.

**Table 1.** Detection performance of the stCSF-equipped observer vs. browsing speed for dataset A (lesion stacks have single micro-calcification), at SSRs of 7, and 14 pixel/deg. The entries corresponding to AUC peaks are in bold face.

| Browsing speed (slice/sec) | AUC ± Std. dev | |
|---|---|---|
| | SSR = 14 pixel/deg | SSR = 7 pixel/deg |
| 1.0 | 0.801136±0.019727 | 0.800228±0.019557 |
| 5.0 | 0.801626±0.019753 | 0.802289±0.019559 |
| 10.0 | 0.802658±0.019952 | 0.807687±0.019648 |
| **15.0** | **0.802850±0.020287** | 0.813819±0.019703 |
| 20.0 | 0.800815±0.020796 | 0.818257±0.019871 |
| **25.0** | 0.795700±0.021290 | **0.819280±0.020180** |
| 30.0 | 0.786696±0.021790 | 0.816057±0.020686 |
| 35.0 | 0.773945±0.022226 | 0.808437±0.021283 |
| 40.0 | 0.758183±0.022564 | 0.796812±0.021857 |
| 45.0 | 0.739797±0.022615 | 0.781314±0.022261 |

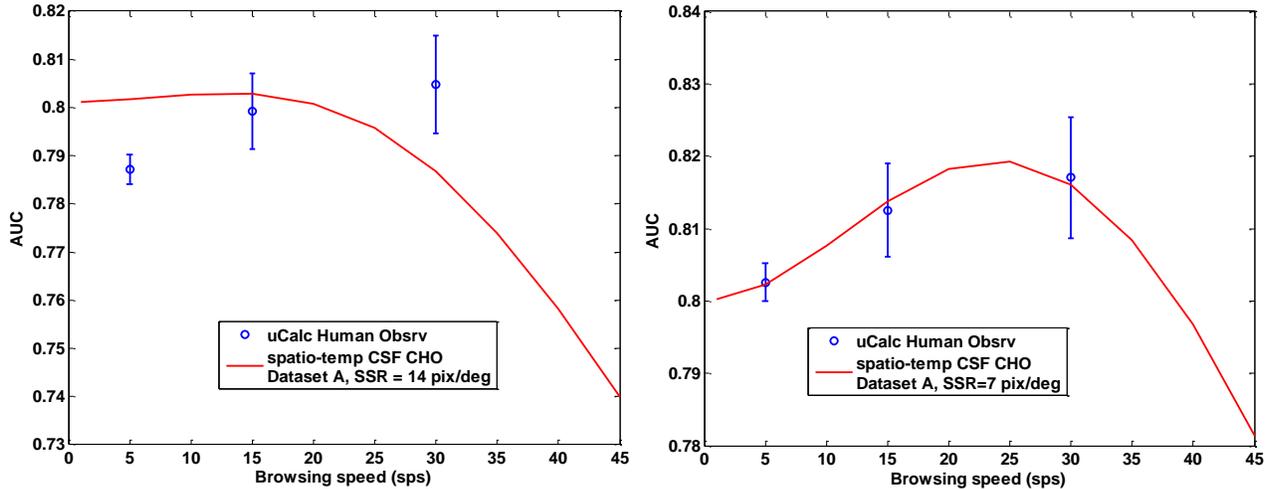

**Figure 5.** *Right*: AUC vs. browsing speed for dataset A at SSR of 7 pixel/deg. Blue samples (with error bars) are from a prior human observer study for detection of micro-calcifications [3]. The red curve is the detection performance of the observer with stCSF. *Left*: same as right, for SSR of 14 pixel/deg.

For dataset B, first, we used micro-calcifications (8-pixel diameter) at a given density to generate the lesion stacks. The detection performance for a variety of browsing speeds with SSR fixed at 7 and 14 pixel/deg is listed in Table 2. The peak behavior is also present in these results, however the peak location depends on SSR. By comparing the two columns of Table 2, it is observed that the peak detection performance is lower for the higher SSR. The same phenomenon is observed in Table 1. The fact that the post-CSF stacks lose details (are further blurred) at a higher SSR explains this observation. Note that a direct comparison of AUC values in Table 1 and Table 2 is not meaningful since the insertion densities are different for the two datasets.

In Figure 6, the results listed in Table 2 are shown together with the human observer results for micro-calcifications in [3], using the method explained in the first paragraph of this section.

**Table 2.** Detection performance of the stCSF-equipped observer vs. browsing speed for dataset B with micro-calcification lesions, at SSRs of 7 and 14 pixel/deg. The entries corresponding to AUC peaks are in bold face.

| Browsing speed (slice/sec) | AUC ± Std. dev | |
|---|---|---|
| | SSR = 14 pixel/deg | SSR = 7 pixel/deg |
| 1.0 | 0.678202±0.011501 | 0.658812±0.010357 |
| 5.0 | 0.679705±0.011405 | 0.662501±0.010186 |
| 10.0 | 0.683514±0.011235 | 0.673050±0.009743 |
| 15.0 | 0.688503±0.011115 | 0.686745±0.009816 |
| 20.0 | 0.692695±0.011117 | 0.700289±0.010231 |
| 25.0 | 0.695320±0.011244 | 0.711045±0.010639 |
| **30.0** | **0.695866±0.011600** | 0.719160±0.010892 |
| 35.0 | 0.694821±0.011977 | 0.723778±0.010932 |
| **40.0** | 0.691011±0.012683 | **0.725776±0.010840** |
| 45.0 | 0.684985±0.013843 | 0.724792±0.010753 |

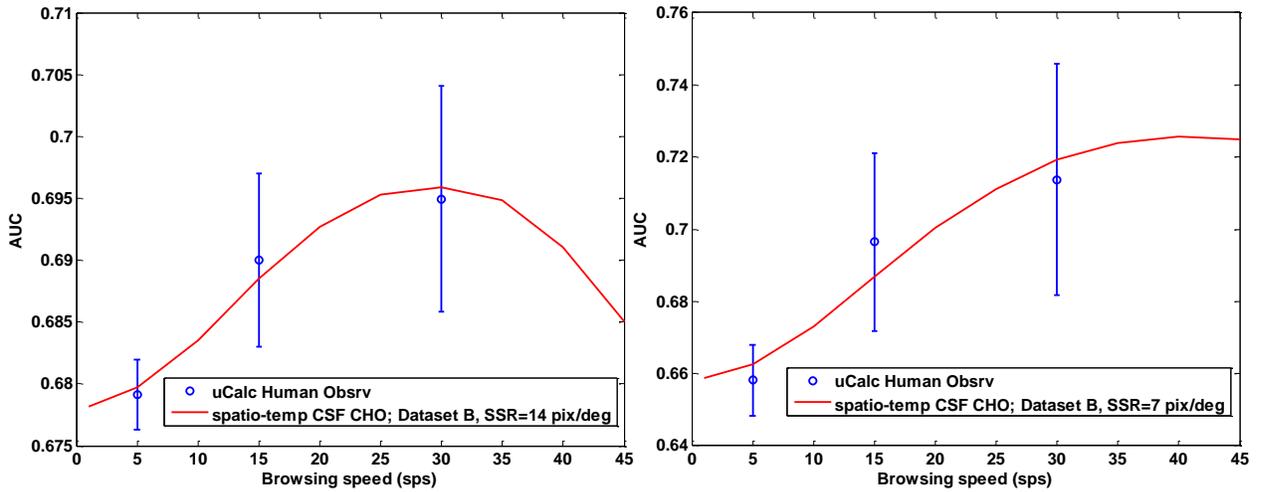

**Figure 6.** *Left*: AUC vs. browsing speed for dataset B (with micro-calcifications lesions) at SSR of 14 pixel/deg. Blue samples (with error bars) are from a prior human observer study for detection of micro-calcifications [3]. The red curve is the detection performance of the observer with stCSF. *Right*: same as left, for SSR of 7 pixel/deg.

Next, for dataset B, we inserted masses (40-pixel diameter) at a given density to generate the lesion stacks. The detection performance for a variety of browsing speeds with SSRs of 7 and 14 pixel/deg is listed in Table 3.

Figure 7 shows simulation results overlaid on those of human observer study for masses in [3], using the method explained in the first paragraph of this section.

## 3.2 Spatial sampling rate

SSR depends on the display's pixel pitch and the viewing distance, assuming orthogonal viewing. Our simulations, at fixed 25 slice/sec browsing speed, show increasing AUC with decreasing SSR (i.e., fewer pixel fit in the field of view) for dataset A. The results are shown in Table 4. This is not consistent with the facts that with decreasing SSR, resulted from smaller viewing distance or larger pixel pitch, (i) more information-bearing pixels fall out of the fovea, and that (ii) the pixel structure of the display becomes more visible: the dark areas in between the light emitting elements of each pixel make up a high frequency noise which is more noticeable at a smaller SSR.

**Table 3.** Detection performance of the stCSF-equipped observer vs. browsing speed for dataset B with mass lesions, at SSRs of 7 and 14 pixel/deg. The entries corresponding to AUC peaks are in bold face.

| Browsing speed (slice/sec) | AUC ± Std. dev | |
|---|---|---|
| | SSR = 14 pixel/deg | SSR = 7 pixel/deg |
| 5.0 | 0.501396±0.011850 | 0.503261±0.011320 |
| 10.0 | 0.516087±0.010884 | 0.525944±0.010206 |
| 15.0 | 0.542477±0.012185 | 0.569007±0.013917 |
| 22.5 | 0.597940±0.018284 | 0.654170±0.021121 |
| 30.0 | 0.647818±0.023345 | 0.723946±0.025662 |
| 37.5 | 0.678916±0.027074 | 0.766593±0.027251 |
| **45.0** | **0.691810±0.029061** | 0.785391±0.028262 |
| **52.5** | 0.690725±0.029888 | **0.788317±0.029643** |
| 60.0 | 0.678989±0.030235 | 0.780556±0.031026 |
| 67.5 | 0.662165±0.029962 | 0.766221±0.031799 |

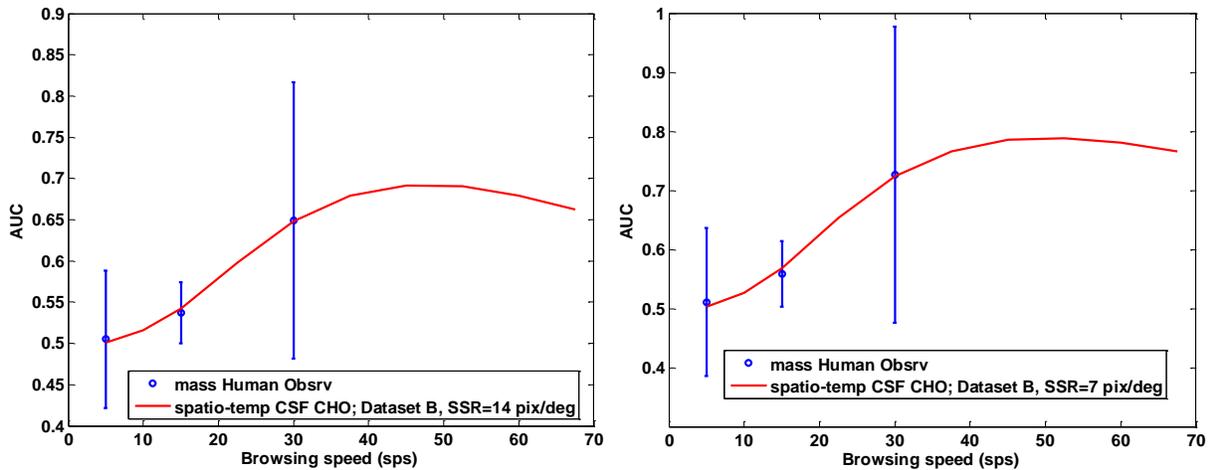

**Figure 7.** *Left*: AUC vs. browsing speed for dataset B (lesioned with masses) at SSR of 14 pixel/deg. Blue samples (with error bars) are from a prior human observer study for detection of masses [3]. The red curve is the detection performance of the observer with stCSF. *Right*: same as left, for SSR of 7 pixel/deg.

Next, we added a model of foveal vision (Section 2.6) to the simulation pipeline. Doing this did not change the trend (i.e., lower SSR still yields higher AUC). As an example, results for soft-thresholding foveal vision are listed in Table 5. Adding a model of pixel structure may change this trend to what we expect: low AUC in low and high SSRs, with an AUC peak in mid SSRs. Simulation of pixel structure, however, is computationally intensive and/or requires a model observer whose results can be correctly compared for input stacks of different sizes. See Section 5 for an idea to fix this.

**Table 4.** Detection performance of the stCSF-equipped observer in a variety of SSRs at browsing speed of 25 slice/sec for dataset A.

| SSR | AUC ± Std. dev |
|---|---|
| 1 | 0.887685±0.013372 |
| 3 | 0.848867±0.017842 |
| 5 | 0.830943±0.019361 |
| 7 | 0.819280±0.020180 |
| 9 | 0.810603±0.020680 |
| 11 | 0.803832±0.020989 |
| 13 | 0.798180±0.021213 |
| 15 | 0.793396±0.021342 |
| 17 | 0.789288±0.021401 |
| 25 | 0.777156±0.021774 |
| 50 | 0.760739±0.022043 |
| 99 | 0.750934±0.021335 |

**Table 5.** Detection performance of the stCSF-equipped observer with foveal vision (soft thresholding) in a variety of SSRs at browsing speed of 25 slice/sec for dataset A.

| SSR | AUC ± Std. dev |
|---|---|
| 0.1 | 0.931626±0.007479 |
| 0.3 | 0.912839±0.009994 |
| 0.5 | 0.899483±0.012337 |
| 0.8 | 0.833326±0.016207 |
| 1 | 0.837633±0.016761 |
| 3 | 0.818672±0.021235 |
| 5 | 0.746200±0.016430 |
| 8 | 0.758987±0.015210 |
| 10 | 0.773042±0.015163 |
| 30 | 0.708260±0.016813 |
| 50 | 0.736531±0.020272 |
| 80 | 0.736547±0.019925 |
| 100 | 0.743223±0.020313 |
| 200 | 0.739457±0.039691 |
| 500 | 0.510106±0.012306 |

### 3.3 Luminance and contrast

The detection performance of simulated displays with 50% less $L_{max}$ (same contrast) and 50% less $L_{max}/L_{min}$ (same $L_{max}$) does not differ significantly with that of the base display. Section 5 gives an idea for fixing this issue.

## 4. CONCLUSIONS

A validated implementation of Barten's stCSF of HVS was integrated with an msCHO. Detection performance results from simulation of DBT stack browsing in time using this observer conform to those of an earlier human observer study [3]. Our results show that AUC peaks at some browsing speed for micro-calcification or mass detection (signal location, if present, is known exactly). The effects of changing display luminance ($L_{max}$) and contrast ($L_{max}/L_{min}$), and SSR (e.g., due to changes in pixel pitch or viewing distance) are also simulated.

## 5. FUTURE WORK

The following are the main avenues that we pursue in continuation of this work.

The effect of changes in display luminance and contrast is perhaps better modeled using a double-ended model observer [11, 12], where HVS's contrast masking mechanism can be modeled. To that end, an image quality metric such as SSIM (Structural Similarity index) [5], which already supports contrast masking, may be used in model observer design.

We learned that modeling foveal vision (Section 2.6) alone cannot generate the expected behavior (Section 3.2). To improve our simulation pipeline, we are working on a computationally efficient modeling of display (sub-)pixel structure. Such a pipeline will produce a larger output (*perceived*) stack when SSR is small so that it can contain more details. The model observer to be used with such a pipeline should be either parameter-free or adaptively change its parameter to generate AUC values that can be correctly compared from one SSR to another. Temporal counterparts of these provisions should be also considered.

## ACKNOWLEDGEMENT

This work is supported by the US National Institutes of Health (R01 grant #CA154444). The authors would like to thank Subok Park and Ljiljana Platiša for their support.